\documentclass[conference]{IEEEtran}
\IEEEoverridecommandlockouts

\usepackage{cite}
\usepackage{amsmath,amssymb,amsfonts}
\usepackage{algorithmic}
\usepackage{graphicx}
\usepackage{textcomp}
\usepackage{xcolor}
\usepackage{fancyhdr}
\usepackage{amsmath}
\usepackage{amsmath,amssymb,amsfonts}
\usepackage{amsthm}
\usepackage{stfloats}
\theoremstyle{definition}
\newtheorem{remark}{\textbf{Remark}}
\def\BibTeX{{\rm B\kern-.05em{\sc i\kern-.025em b}\kern-.08em
    T\kern-.1667em\lower.7ex\hbox{E}\kern-.125emX}}
\DeclareMathAlphabet{\mathpzc}{OT1}{pzc}{m}{it}
\begin{document}



\title{NightSight: Passive Computation for Navigation in Dark Using Events

\thanks{}
}

\author{
\IEEEauthorblockN{Deepak Singh*\textsuperscript{1}, Brijan Vaghasiya*\textsuperscript{1}, Shreyas Khobragade*\textsuperscript{1}, Nitin J Sanket\textsuperscript{1}}

\IEEEauthorblockA{\textsuperscript{1}\textit{Perception and Autonomous Robotics (PeAR) Group} \\
\textit{Worcester Polytechnic Institute} \\
Worcester, USA \\
\{dsingh1, bcvaghasiya, skhobragade, nitin\}@wpi.edu}

\thanks{*Equal Contribution. Author order decided randomly.}
}

\maketitle

\fancypagestyle{withfooter}{
\renewcommand{\headrulewidth}{0pt}
\fancyfoot[C]{\footnotesize Accepted to the Challenges and Opportunities of Neuromorphic Field Robotics and Automation IEEE ICRA Workshop - 2026}
}
\thispagestyle{withfooter}
\pagestyle{withfooter}


\begin{abstract}
Small aerial robots are particularly well-suited for search and rescue in confined and hazardous environments due to their agility, low cost, and ability to traverse through cluttered spaces that are inaccessible to larger platforms. However, enabling autonomous navigation in complete darkness remains a significant challenge, because small aerial robots cannot easily accommodate perception systems that demand substantial payload, power, or computation. In this work, we present a lightweight perception approach that combines a monocular event camera, a coded aperture lens, and an infrared dot projector to enable navigation in such conditions. The projected pattern, when imaged through the coded aperture, produces depth dependent blur signatures that implicitly encode scene geometry. We train a convolutional neural network to decode these signatures into dense depth maps using only synthetic data generated from a simple planar wall setup. Despite this minimal training regime, the model generalizes zero-shot to complex real-world scenes. Our system operates in real time at 20 Hz on a NVIDIA Jetson Orin Nano$^\text{TM}$, demonstrating suitability for resource-constrained platforms. We further analyze the impact of different coded aperture designs on depth estimation performance. Our approach gives high accuracy ($l_1 \ \text{error} \ 7.0cm$) upto 2.5m range ($2.80\% \ \text{error}$).  These results highlight the potential of combining structured illumination, coded optics, and event-based sensing for enabling robust perception and navigation in complete darkness.
\end{abstract}


\section{Introduction}

Search and rescue operations following disasters such as earthquakes or storms are often conducted in complete darkness due to widespread power outages. Aerial robots are well suited for such scenarios, as they can rapidly explore collapsed structures and cluttered environments that are inaccessible to ground responders. However, navigating through these tight spaces requires very small aerial robots. The limited size of such platforms severely restricts the sensors, computation, and power they can carry, making perception in dark environments a fundamental challenge.

\begin{figure}[t!]
    \centering
    \includegraphics[width=\linewidth]{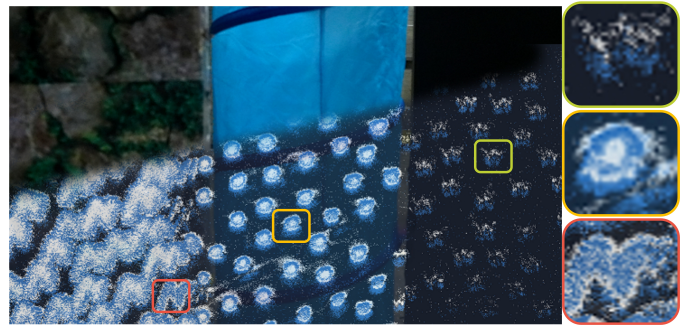}
\caption{\textit{NightSight} estimates dense metric depth in completely dark scenes using an event camera equipped with a coded aperture lens and a structured infrared illumination source. The method generalizes zero-shot to real-world environments containing obstacles of varying sizes, shapes, and textures. On the left, the highlighted patterns show how the projected dot pattern exhibits depth-dependent blur, with the coded-aperture signature changing in size as depth varies (red to green indicates increasing depth).}
    \label{fig:banner}
\end{figure}

Existing approaches typically rely on heavy sensing systems or high-power illumination. For instance, several state-of-the-art systems employ LiDARs or cameras paired with powerful lighting sources ($>$10W) to perceive the environment in darkness \cite{Barfoot_cam_lidar, radar_cam_fusion, kostas_subt}. Other approaches combine multiple sensing modalities such as thermal cameras \cite{thermal}.  While such solutions are feasible for larger robots, the associated payload and power requirements make them unsuitable for small aerial or ground robots \cite{sanket2020prgflowbenchmarkingswapawareunified}. A related approach, AsterNav \cite{asternav}, proposed a computationally efficient strategy for depth perception in darkness using structured illumination and coded optics. However, it relies on a conventional frame-based camera, whose limited dynamic range restricts the usable sensing distance in low-light conditions, while also affecting the depth quality.

These constraints motivate revisiting how perception is achieved in biological systems. Biological sensing follows a principle of parsimony, prioritizing efficiency by extracting only task relevant information, where simple sensing mechanisms exploit physical interactions with the environment to produce signals sufficient for action \cite{animalEyes, ajna}. Instead of relying on extensive computation, living beings encode useful environmental cues directly in the physics of their sensory systems \cite{chameleons_accomodation}. For example, small flying insects such as moths can navigate complex environments under extremely low-light conditions despite possessing very limited neural resources, demonstrating how effective perception can arise from parsimonious sensing strategies \cite{moth}. A similar principle was used in \cite{blurringForClarity}, which used depth from defocus with event cameras to perceive scene depth, with minimal computational requirements.

Inspired by this principle, we propose a monocular depth sensing approach that combines an event camera with custom optical elements to enable depth perception in complete darkness. Event cameras provide significantly higher dynamic range than conventional frame based cameras and capture brightness changes with high temporal resolution, enabling reliable perception under extremely low-light  \cite{tobi_events}. Although defocus-based depth cues have been widely investigated for frame-based imaging systems, their application to event cameras remains largely unexplored. \cite{ryad} proposed one of the first works for depth from defocus in events using a Spiking Neural Network based approach, with a varifocal liquid lens atop an event camera. While focus can be readily quantified using image gradients in frame-based cameras, this is not straightforward for event data due to their asynchronous sensing nature. \cite{lin2022autofocus} proposed the use of event rate as focus measure, and used that to autofocus an event camera. Depth from defocus in events was further utilized in \cite{eventFocalStack}, \cite{xue2025event}, where a focal stack was built for events generated from different depths, which was then processed by a deep learning model to estimate depth. In contrast to the methods used in these works, our system projects a sparse structured dot pattern onto the scene and images it through a coded aperture integrated into the lens. The coded aperture induces depth-dependent blur patterns, causing each projected dot to appear as a scaled version of the aperture shape whose size varies with scene depth (as shown in Fig. \ref{fig:banner}). These depth dependent optical patterns provide strong physics based cues that can be decoded by a learning based model. Importantly, the coded aperture introduces these cues \emph{passively}, requiring no additional electrical power. We therefore refer to this strategy as \emph{passive computation}, where part of the perception pipeline is implemented directly in optics rather than digital processing \cite{blurringForClarity}.

\begin{figure}[!t]
    \centering
    \includegraphics[width=\linewidth]{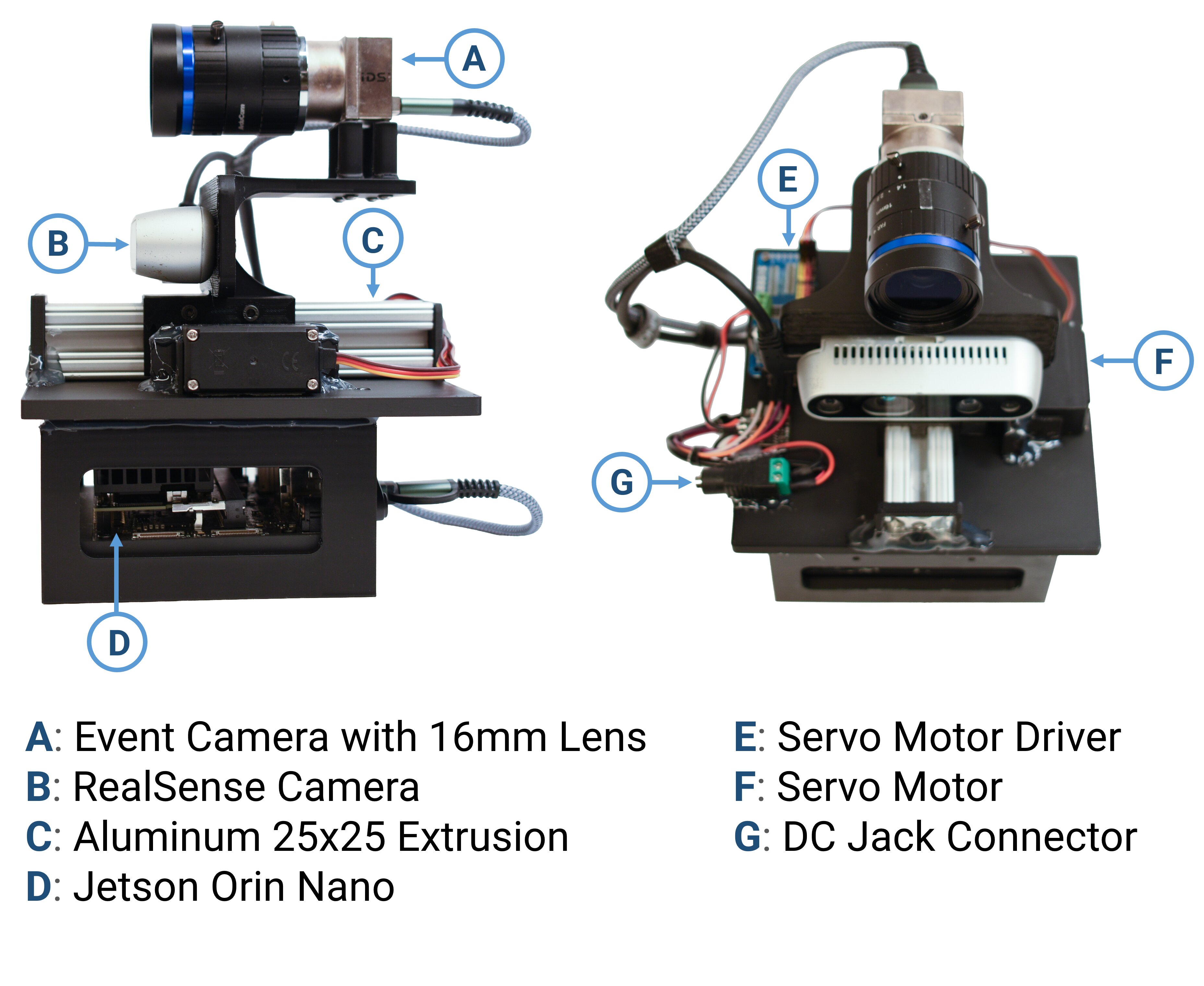}
    \caption{Experimental data collection setup featuring an event camera and a RealSense depth sensor mounted on an aluminum extrusion rail. A servo-actuated rack-and-pinion mechanism induces controlled angular motion ($\approx$10° at maximum speed) to emulate vibration during acquisition. This setup enables synchronized capture of event and depth data under dynamic conditions for evaluating motion robustness.}
    \label{fig:data_collection_setup}
\end{figure}

Using these cues, we train our convolutional neural model (architecturally similar to the one proposed in \cite{asternav}) to recover dense depth from event data. The model is trained entirely in simulation using data generated by projecting the structured dot pattern onto a planar wall and recording the resulting event streams. To generate motion for event sensing, we use a servo-driven linear actuator setup (Fig. \ref{fig:data_collection_setup}) that quickly moves back and forth, producing brightness changes that simulate small camera movements. Notably, the model is trained only on this simple motion-generated dataset and is never explicitly trained across different velocities. Our approach gives accurate metric depth estimation, with $l_1$ error of just $7cm$ for depth range of $2.5\text{m}$, while running at 20\,Hz on NVIDIA Jetson Orin$^\text{TM}$, enabling accurate real-time onboard depth estimation suitable for autonomous robot navigation.

The main contributions of this work are summarized as follows:

\begin{itemize}
    \item We present a monocular depth sensing approach for aerial robot navigation in complete darkness by combining an event camera, structured illumination, and a coded aperture that encodes depth cues directly in optics via passive computation.

    \item We show zero-shot generalization of our depth estimation approach, by training the model entirely on simulated data and testing in real-world scenes.
    
    \item We benchmark multiple coded aperture patterns and evaluate their effectiveness for depth estimation using event-based sensing.
\end{itemize}


\section{Passive Computation using Coded Aperture and Defocus}
\emph{Passive Computation} as defined in \cite{blurringForClarity, asternav} refers to an approach wherein the optical components consume no active electrical power, and process the input signal passively in the physical domain, thereby reducing the computational burden onboard. \cite{blurringForClarity} first introduced the concept of passive computation for robot perception using event cameras by exploiting depth-dependent defocus blur induced by a fully open aperture. \cite{asternav} later extended this idea by employing a coded aperture to encode richer depth-dependent blur patterns in the observed image, enabling navigation in dark environments using a frame-based camera. Our work further develops this paradigm by combining an event camera with a coded aperture lens and a structured lighting source to enable depth perception using blur patterns in complete darkness. 

For a lens with focal length $f$ and aperture number $N$, the amount of defocus blur varies with the distance of the scene point from the camera. If the lens is focused at distance $Z_f$ (focus distance), the size of the blur circle for a point located at distance $Z$ from the camera is given by
\begin{equation}
    s = \frac{f^2}{N}\frac{|Z-Z_f|}{Z (Z_f - f)}
    \label{eq:blur_circle}
\end{equation}

\begin{figure}[t!]
    \centering
    \includegraphics[width=\linewidth]{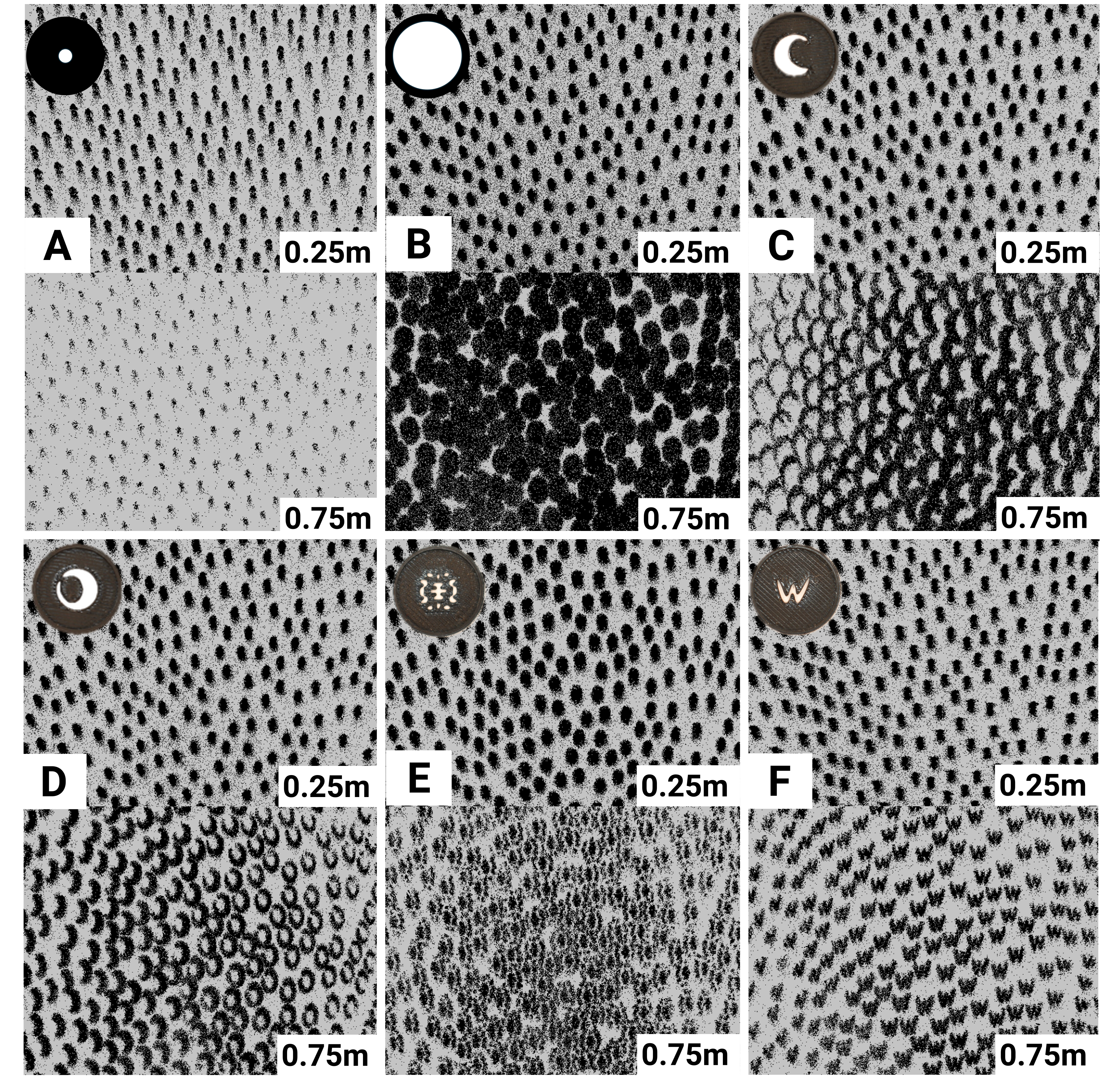}
    \caption{Visual comparison of pinhole (A), fully open (B), and coded apertures ($\text{Zhou}_1$ (C), $\text{Zhou}_2$ (D), $\text{Levin}$ (E) and $W$ aperture (F)) imaging at object distances of 0.25 m (top row) and 0.75 m (bottom row). The pinhole aperture preserves near-uniform sharpness across depths due to its extended depth of field, while the fully open aperture introduces significant defocus blur as the scene deviates from the focal plane. In contrast, the coded aperture produces structured, depth-dependent blur patterns that vary more distinctly with defocus. Notably, the W-shaped coded aperture demonstrates the lowest reconstruction error among all aperture designs, particularly at depths farther from the focal plane, indicating its superior capability to encode discriminative defocus cues for robust depth estimation.}
    \label{fig:psf_patterns}
\end{figure}

The corresponding image generated by the camera (denoted by $I$) is given as 
\begin{equation}
    I = I_o \circledast h(Z, Z_f)
    \label{eq:psf}
\end{equation}
where $I_o$ is all-in-focus image (or completely sharp image), $h(Z, Z_f)$ is the point spread function (PSF) of the camera, whose effect on the image is dependent on the object depth ($Z$) and camera focus distance ($Z_f$), and $ \circledast$ is the convolution operator. This PSF causes the depth dependent blurring in the image and acts as a low pass filter, with the amount of blurring dependent on blur circle size given in Eq. \ref{eq:blur_circle}. Fig. \ref{fig:psf_patterns} shows how the IR pattern projected on a plane wall is imaged through a coded aperture camera. When at focus (0.25m in Fig. \ref{fig:psf_patterns}), the image contains sharp circular patterns, but when moved away, the blur pattern shape changes to that of the aperture opening. For an event camera, $I$ and $I_o$ are the latent intensity images.

While depth from defocus works have mostly presumed circular apertures \cite{pentland1987depthoffield, subbarao1994depth} , using a coded aperture induces more distinguishable patterns whose variation with depth is more pronounced, and hence easier to estimate depth \cite{levin, nayar}. The pattern of zero-crossings in the Fourier domain of a coded aperture is different than that of circular apertures, making it more depth sensitive \cite{nayar}. 

\subsection{Depth from Defocus (DfD) in Events}
Event cameras, also called silicon retinas, sense change in light intensity, instead of recording absolute intensity values \cite{tobi_events}. These cameras operate asynchronously, with each pixel recording data independently of others. Events are recorded at a pixel when log difference of intensity crosses a threshold \cite{tobi_events, gehrig2024lowlatency}. For a scene with constant lighting, the rate of change of intensity is
\begin{equation}
\frac{dI}{dt} = \frac{\partial I}{\partial x} \frac{dx}{dt} + \frac{\partial I}{\partial y}\frac{dy}{dt} + \frac{\partial I}{\partial t} = \dot{x}\frac{\partial I}{\partial x} + \dot{y}\frac{\partial I}{\partial y}
\label{eq:event_velocity_equation}
\end{equation}
where $\frac{\partial I}{\partial t} = 0$ (constant scene lighting assumption). Here, $\dot{x}$ and $\dot{y}$ are pixel velocities in the direction of movement. Eq. \ref{eq:event_velocity_equation} can be written as, 
\begin{equation}
\frac{dI}{dt} = \mathbf{v} \cdot \nabla_{\mathbf{x}} I 
\label{eq:dot_product}
\end{equation}
here $\nabla_{\mathbf{x}} I = \left[\frac{\partial I}{\partial x}, \frac{\partial I}{\partial y}\right]^T$ represents the spatial image gradient, quantifying the rate of intensity change in the $x$ and $y$ directions. The term $\mathbf{v} = (\dot{x}, \dot{y})$ denotes the pixel velocity or optical flow (which depends on robot velocity $\mathbf{V} = (V_x, V_y, V_z)$ and scene depth $Z$), describing the apparent motion of pixels on the image plane. Eq. \ref{eq:dot_product} shows that rate of change intensity or rate of events, vary with velocity, depth and gradient in latent intensity image ($I$). 

Since event generation depends on gradient of latent image intensity, sharp regions generate more events than blurred regions. Eq. \ref{eq:psf} shows how imaging happens in a frame based camera, given the PSF of the camera. Since events are generated due to change in intensity, Eq. \ref{eq:psf} can be adapted for events as follows
\begin{equation}
    \nabla I = \nabla I_o \circledast h(Z, Z_f)
\label{eq:events_psf}
\end{equation}
Therefore, for events, the PSF reduces the strength of edges and consequently the overall events. Hence, the quality of depth estimation depends upon amount of depth dependent blurring of the scene, i.e. how strongly the PSF blurs out objects at different depths. Here we presume that there is nothing between the robot and the focus distance $Z_f$, i.e., the obstacles in the scene are always at distance $Z > Z_f$ from the camera. The strength of PSF blurring at $Z$, depends on blur circle size at that depth, which from Eq. \ref{eq:blur_circle} depends on lens focal length $f$, and aperture number $N$ (for a fixed $Z_f$). This helps us select the ideal focal length and aperture number to get accurate dense metric depth maps. Also, in this work, since we are projecting a structured lighting pattern on the obstacles, the texture remains constant, as scene illumination is $<$ 1 milliLux. The various factors affecting the depth estimation quality and factors governing the choice of optical and event camera parameters are summarized in remarks below.

\begin{remark}
\label{remark:1}
\textit{For a robot moving at constant speed, in a dark scene, illuminated by an external structured lighting source, areas in the scene which are closer to $Z_f$ generate more events as compared to farther ones.}
\end{remark}

\begin{remark}
\label{remark:2}
\textit{Depth estimation quality varies with varying speed, as the event count also varies. However, as long as the product of robot velocity and event accumulation time is constant, depth prediction is consistent.}
\end{remark}

\begin{remark}
\label{remark:3}
\textit{Depth sensitivity varies with aperture opening ($N$) and focal length $f$. Changing these affects the sensitivity and thus range till which depth can be estimated. However, depth sensitivity does not depend on $Z_f$, for cases where $Z_f >> f$.}
\end{remark}

From Eq. \ref{eq:events_psf}, the gradients in latent image or intensity change depends upon the PSF $h(Z, Z_f)$ or consequently the blur size (Eq. \ref{eq:blur_circle}). Thus, the maximum depth range and depth sensitivity depends on the variation of blur with depth $\frac{ds}{dZ}$ as,
\begin{equation}
    \frac{ds}{dZ} = \frac{f^{2} Z_f}{N (Z_f - f) Z^{2}}
\label{eq:blur_derivative}
\end{equation}

For $Z_f >> f$, Eq. \ref{eq:blur_derivative} becomes
\begin{equation}
    \frac{ds}{dZ} = \frac{f^{2}}{N Z^{2}}
\label{eq:blur_derivative_zf>f}
\end{equation}

\begin{figure}[t!]
    \centering
    \includegraphics[width=\linewidth]{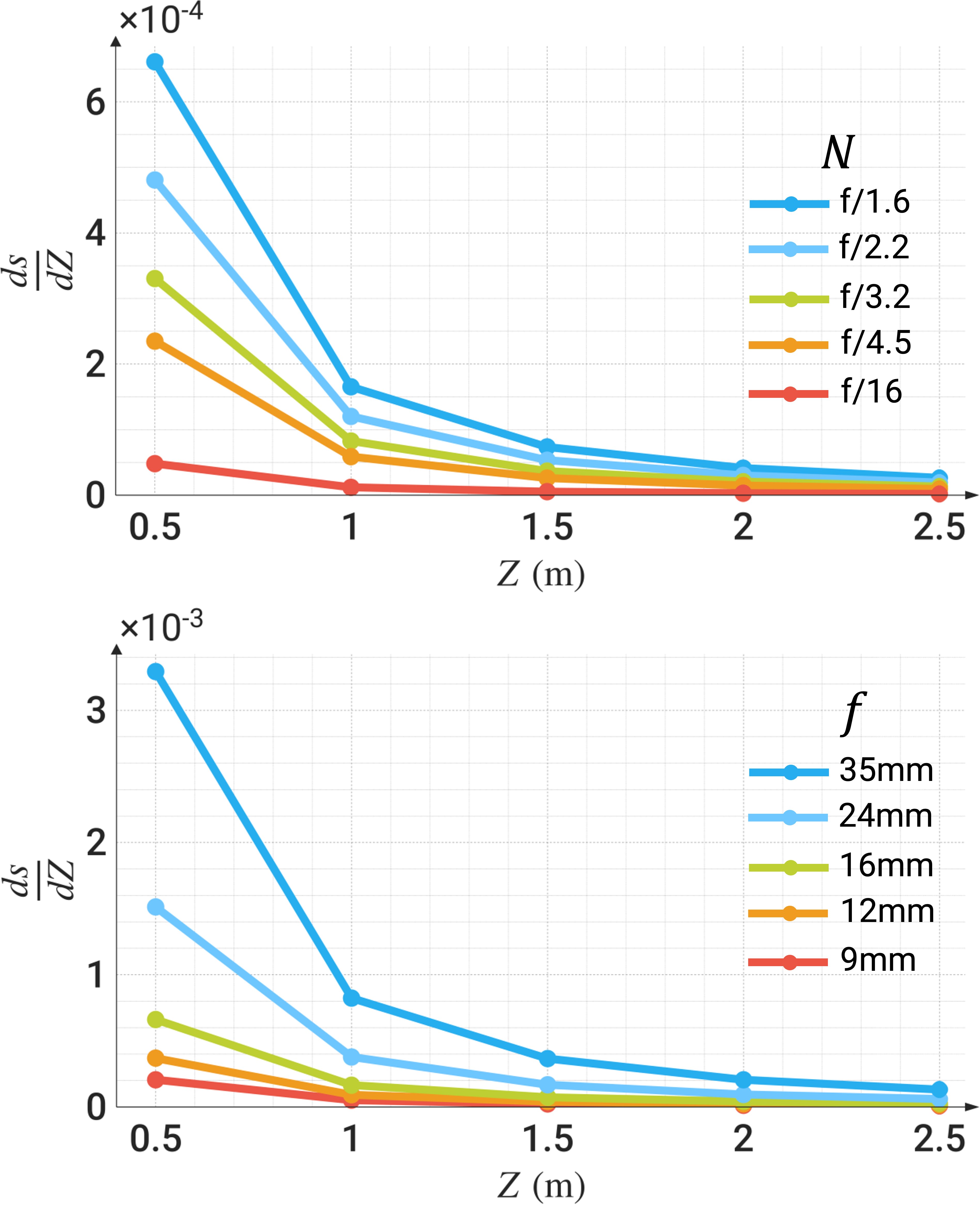}
    \caption{Variation of $\frac{ds}{dZ}$ with object distance ($Z$), with varying aperture numbers $N$ (top), and focal length $f$ (bottom). It can be seen, as aperture opening reduces, or focal length reduces (increasing field of view), the rate of change of blur with distance falls, thus limiting the depth sensing capability using defocus cues. For both the cases, $Z_f$ was kept as 0.50 m. For the top part, $f$ was set to 35mm, and for varying focal length, $N$ was set to $f/16$.}
    \label{fig:plots}
\end{figure}

As shown in Eq. \ref{eq:blur_derivative}, blur sensitivity reduces as $Z$ increases, therefore the maximum sensing range also decreases (see Fig.~\ref{fig:plots}). However, this sensitivity and maximum range can be increased by changing $f$, $N$ and $Z_f$ (as long as it is comparable to $f$).

\subsection{Training Data Generation}
Our training data generation pipeline follows similar approach to \cite{asternav}, with the difference that we now acquire events over an accumulation time, which requires the camera to move. We built a linear actuator mechanism driven by a servo motor that oscillates the camera back and forth about a point, thereby generating events(Fig. \ref{fig:data_collection_setup}). At any time $t$, we capture event volume ($\mathcal{E}$ = {$(\textbf{x}, t, p)_{k}$}, where p is polarity and k is event index) for $\Delta T$ accumulation time. Inspired from \cite{blurringForClarity} we construct event frames $\mathpzc{E}$ from the event volume by taking average of total event occurrences per pixel. Similar to \cite{blurringForClarity, evdodgenet}, we treat both event polarities equally, resulting in a single 2D event frame $\mathpzc{E}$. Once these event frames are generated, we create synthetic training dataset for our network using the Point Spread Function (PSF) patterns obtained at different depths. Unlike \cite{asternav}, blank images are used as background in the dataset generation pipeline.

\section{Experiments and Results}

Our experimental setup uses an IDS uEye XCP-E event camera with a $1/2.3"$ sensor and a $16\,\mathrm{mm}$ $f/1.4$ lens. For training and testing models, we set accumulation time $\Delta T$ to 50ms. Different coded aperture patterns are 3D-printed and mounted at the rear of the lens. Each aperture has a diameter of $9\,\mathrm{mm}$ and a thickness of $0.3\,\mathrm{mm}$, and is fabricated using a Bambu Lab X1 Carbon printer with black eSUN PLA+ filament. For the setup shown in Fig. \ref{fig:data_collection_setup}, we use a TowerPro SG-5010 servo motor for moving the camera.

We compare several aperture configurations, including a pinhole $(f/16)$, a fully open aperture $(f/1.6)$, Levin's aperture~\cite{levin}, two apertures from Zhou et al.~\cite{nayar}, denoted as $\text{Zhou}_1$ and $\text{Zhou}_2$, and a $W$-shaped aperture inspired by cuttlefish eyes~\cite{cuttlefish}.

\begin{figure}[h]
    \centering
    \includegraphics[width=\linewidth]{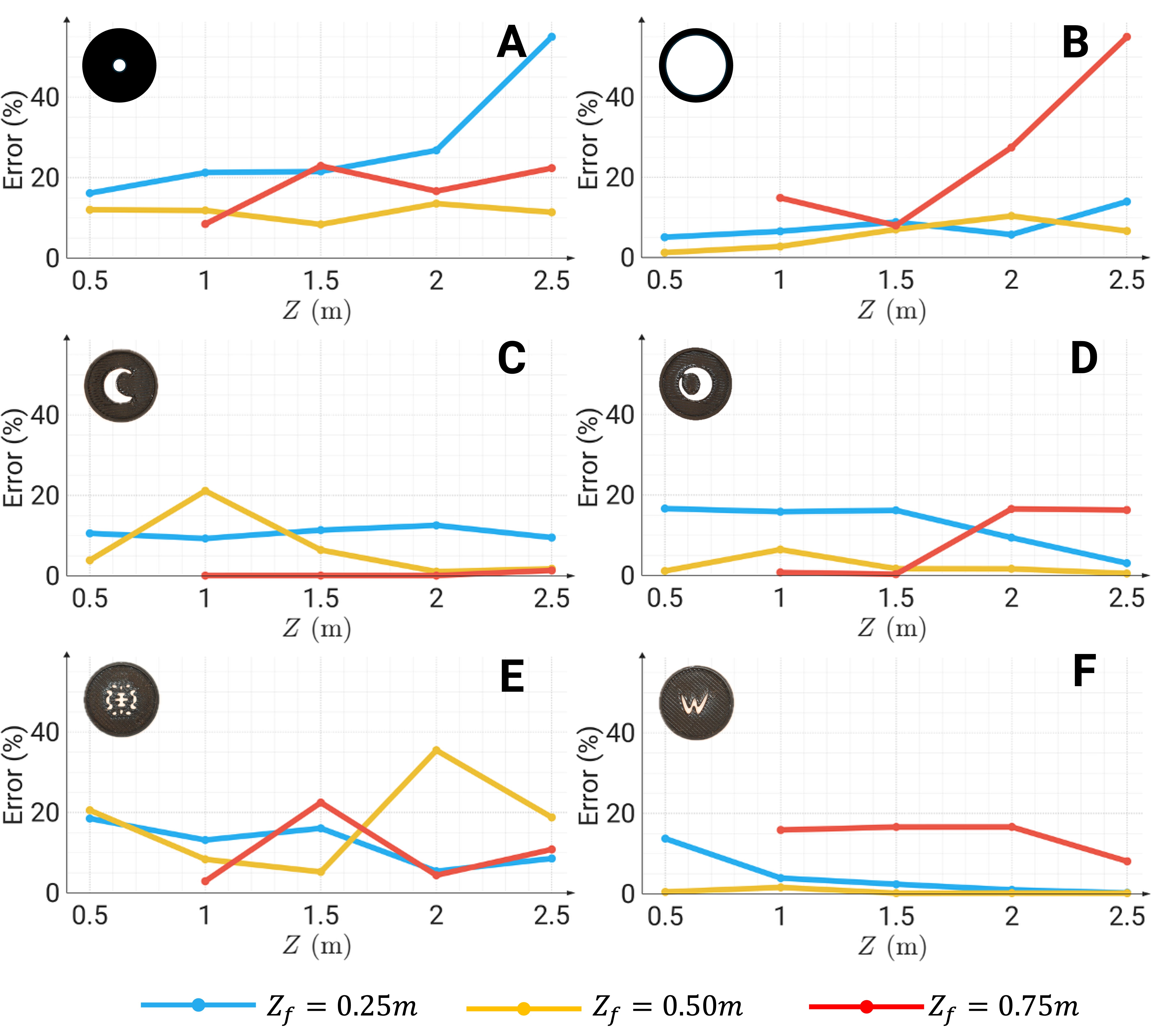}
    \caption{$l_1$ Depth Error with object distance ($Z$) for different focus distances ($Z_f$) for (a) Pinhole (f/16), (b) Fully Open (f/1.4), (c) $Zhou_1$, (d) $Zhou_2$, (e) Levin, and (f) W shaped aperture.}
    \label{fig:error_plots}
\end{figure}

To benchmark depth estimation accuracy, we place a box-shaped obstacle at varying distances from the camera. The obstacle is covered with a rock-and-moss texture and has dimensions ranging from $1.15\,\mathrm{m}$ to $1.28\,\mathrm{m}$. For each aperture, we evaluate performance at focus distances $Z_f \in \{0.25, 0.50, 0.75\}\,\mathrm{m}$. The resulting depth errors across object distances are shown in Fig.~\ref{fig:error_plots}. The best results are obtained with the $W$-shaped aperture at $Z_f = 0.5\,\mathrm{m}$, which achieves an error of $2.80\%$ over a $2.5\,\mathrm{m}$ depth range, corresponding to an $L_1$ error of $7\,\mathrm{cm}$.

\begin{figure*}[!t]
    \centering
    \includegraphics[width=\linewidth]{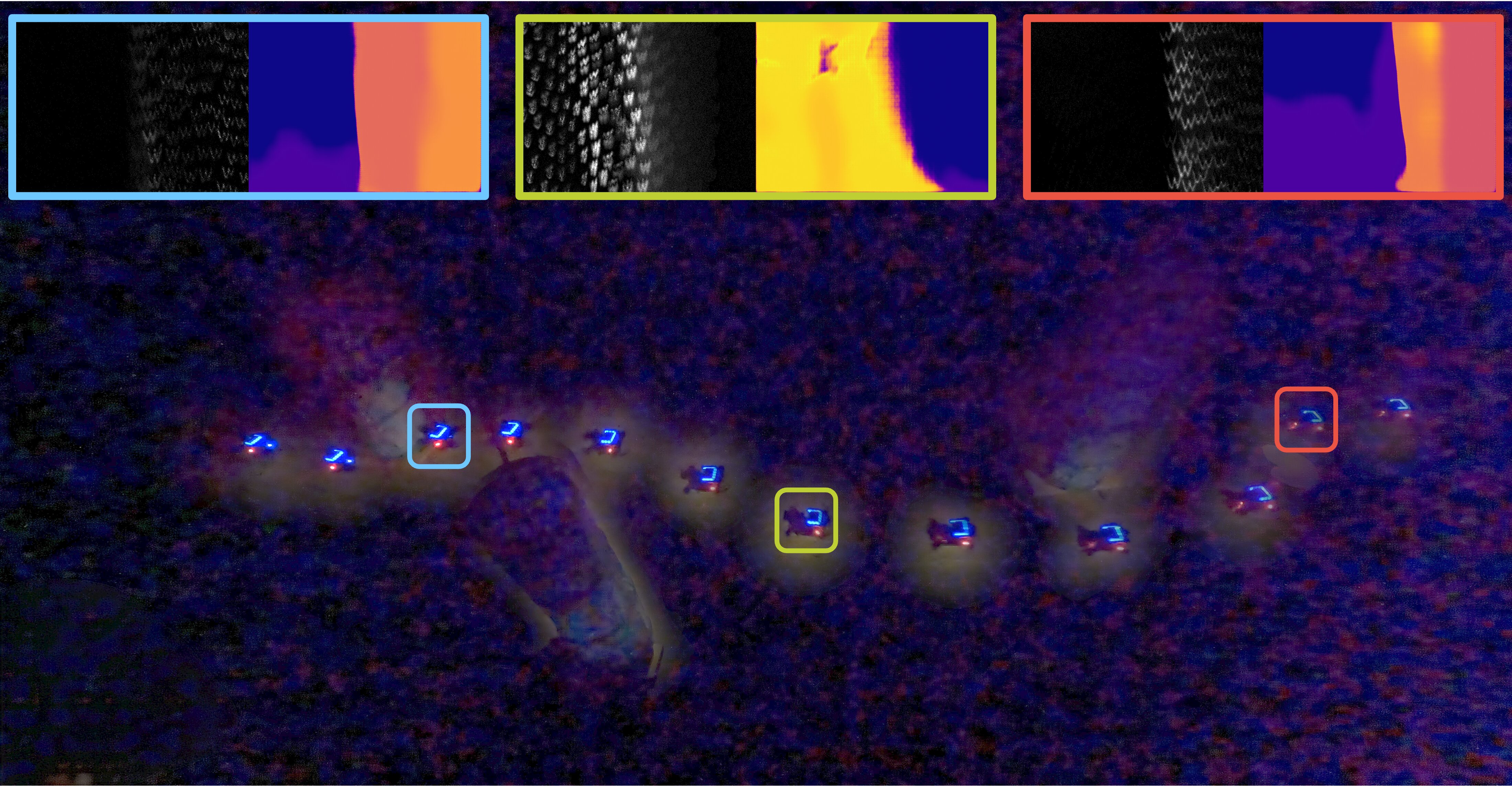}
    \caption{Depth estimation in real-world flight test in the dark. The image shows the trajectory of the drone through a scene with multiple cylindrical obstacles in complete darkness ($<1 \text{milliLux}$). Each colored window shows the input images to our model (on the left) and the predicted depth from the model (on the right) at various instances throughout the trajectory. The robot has a blue colored light on the back and a downward facing light on the bottom, for aiding optical flow in completely dark scenes.}
    \label{fig:hardwareruns}
\end{figure*}

When $Z_f = 0.25\,\mathrm{m}$, the blur is more sensitive to depth changes than at $Z_f = 0.5\,\mathrm{m}$ or $0.75\,\mathrm{m}$, as shown in Eq. \ref{eq:blur_derivative}. Furthermore, for objects closer to the camera, the blur varies more rapidly with depth ($Z$), as indicated by Eq.~\ref{eq:blur_derivative}. This increased sensitivity amplifies the effect of small depth variations and leads to higher prediction errors. We evaluate our approach in a real-world flight test in the dark (Fig.~\ref{fig:hardwareruns}).

\begin{figure}[!h]
    \centering
    \includegraphics[width=\linewidth]{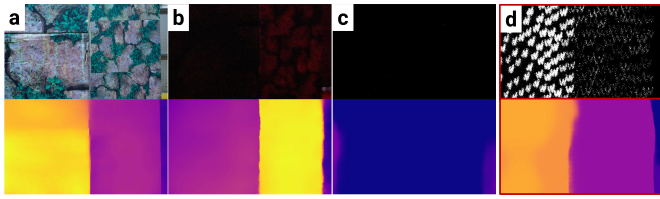}
    \caption{Qualitative comparison of depth estimation using DepthPro and our model. Left to right: (a) D5600 Camera image with scene lighting and the corresponding depth from DepthPro in the image below, (b) D5600 Camera image without scene lighting, (c) Image from coded aperture event camera without the structured lighting pattern, and (d) Our event camera setup with structured lighting. The second row shows the depth prediction.}
    \label{fig:dslr_comparison}
\end{figure}

We further evaluate our approach by comparing it with a state-of-the-art monocular depth estimation model \cite{depthPro} on images captured using a Nikon D5600 DSLR camera with 23.5 mm x 15.6 mm sensor size. Fig.~\ref{fig:dslr_comparison} presents images from the DSLR as well as from our event camera equipped with a coded aperture lens. Fig.~\ref{fig:dslr_comparison}(a) shows the scene under illumination (scene luminosity $155\,\mathrm{lux}$), while Fig.~\ref{fig:dslr_comparison}(b) shows the same scene in near-complete darkness (scene luminosity $1\,\mathrm{milliLux}$). For Fig.~\ref{fig:dslr_comparison}(b), the DSLR camera settings were configured to maximize light capture, with a shutter speed of $1\,\mathrm{s}$, aperture $f/5.6$, and ISO $25600$. We additionally show the response of our event camera without structured illumination in Fig.~\ref{fig:dslr_comparison}(c), and with the structured infrared (IR) lighting source enabled in Fig.~\ref{fig:dslr_comparison}(d), with the corresponding predicted depth maps shown in the second row. As observed, even with a high-end DSLR operating at extreme exposure and ISO settings, the scene is barely perceptible in near-dark conditions, highlighting the limitations of conventional imaging in such environments, which would be even more pronounced for lightweight robotic cameras. Furthermore, despite its high dynamic range, the event camera is unable to perceive the scene without structured IR illumination in contrast, with the addition of structured lighting, our system is able to recover meaningful scene structure and produce accurate depth estimates. 

\section{Conclusion and Future Work}

In this work, we introduced a lightweight perception framework for enabling onboard depth estimation for small aerial robot navigation in complete darkness. By combining a monocular event camera with a coded aperture lens and an infrared dot projector, our system leverages depth-dependent blur patterns as an implicit encoding of scene geometry. Despite being trained solely on synthetic data derived from a simple planar setup, the proposed model demonstrates robust zero-shot generalization to complex real-world environments, while enabling real time inference (20 Hz on a Jetson Orin Nano). Future work will focus towards increasing depth sensing range, 
to enable high speed navigation in cluttered dark scenes for small aerial robots.

\end{document}